\documentclass[journal,twoside,web]{ieeecolor}
\usepackage{generic}
\usepackage{cite}
\usepackage{amsmath,amssymb,amsfonts}
\usepackage{algorithm}
\usepackage{algorithmic}

\usepackage{graphicx}
\usepackage{algorithm,algorithmic}
\usepackage{hyperref}
\hypersetup{hidelinks}
\usepackage{textcomp}

\usepackage{graphicx}
\usepackage{amsmath}
\usepackage{amssymb}
\usepackage{helvet}
\usepackage{courier}
\usepackage{float}
\usepackage{color}
\usepackage{booktabs}
\usepackage{multirow}
\usepackage{bbding}
\usepackage{pifont}
\usepackage{hyperref}
\usepackage[normalem]{ulem}
\usepackage{arydshln}
\usepackage{graphicx}
\usepackage{textcomp}

\usepackage{multirow}
\usepackage{bbm}
\usepackage{graphicx}
\usepackage{amssymb}
\usepackage{helvet}
\usepackage{courier}
\usepackage{float}
\usepackage{color,soul}
\usepackage{hyperref} 
\usepackage{amsmath,amsfonts}
\def\BibTeX{{\rm B\kern-.05em{\sc i\kern-.025em b}\kern-.08em
    T\kern-.1667em\lower.7ex\hbox{E}\kern-.125emX}}
\begin{document}
\title{PSTNet: Enhanced Polyp Segmentation with Multi-scale Alignment and Frequency Domain Integration}
\author{Wenhao Xu, Rongtao Xu, Changwei Wang, Xiuli Li, Shibiao Xu, Li Guo
\thanks{Wenhao Xu, Shibiao Xu and Li Guo are with School of Artificial Intelligence, Beijing University of Posts and Telecommunications, Beijing 100876, China. Rongtao Xu is with the State Key Laboratory of Multimodal Artificial Intelligence Systems, Institute of Automation, Chinese Academy of Sciences, Beijing, China. 
Changwei Wang is with the Key Laboratory of Computing Power Network and Information Security, Ministry of Education, Shandong Computer Science Center (National Supercomputer Center in Jinan), Qilu University of Technology (Shandong Academy of Sciences), Jinan, 250013, China; Shandong Provincial Key Laboratory of Computer Networks, Shandong Fundamental Research Center for Computer Science, Jinan, China; the State Key Laboratory of Multimodal Artificial Intelligence Systems, Institute of Automation, Chinese Academy of Sciences, Beijing, China.
Xiuli Li is with AI Lab, Deepwise Healthcare, Beijing 100080, China.}
\thanks{Shibiao Xu is the corresponding author (shibiaoxu@bupt.edu.cn).}
}

\maketitle

\begin{abstract}
Accurate segmentation of colorectal polyps in colonoscopy images is crucial for effective diagnosis and management of colorectal cancer (CRC). However, current deep learning-based methods primarily rely on fusing RGB information across multiple scales, leading to limitations in accurately identifying polyps due to restricted RGB domain information and challenges in feature misalignment during multi-scale aggregation.
To address these limitations, we propose the Polyp Segmentation Network with Shunted Transformer (PSTNet), a novel approach that integrates both RGB and frequency domain cues present in the images. PSTNet comprises three key modules: the Frequency Characterization Attention Module (FCAM) for extracting frequency cues and capturing polyp characteristics, the Feature Supplementary Alignment Module (FSAM) for aligning semantic information and reducing misalignment noise, and the Cross Perception localization Module (CPM) for synergizing frequency cues with high-level semantics to achieve efficient polyp segmentation.
Extensive experiments on challenging datasets demonstrate PSTNet's significant improvement in polyp segmentation accuracy across various metrics, consistently outperforming state-of-the-art methods. The integration of frequency domain cues and the novel architectural design of PSTNet contribute to advancing computer-assisted polyp segmentation, facilitating more accurate diagnosis and management of CRC.

\end{abstract}


\section{Introduction}
\label{sec:introduction}

\IEEEPARstart{C}{olorectal} cancer (CRC) often begins with the development of epithelial polyps within the colon or rectum, which are considered precursors to this malignant condition. While these polyps are initially non-cancerous, there is a risk that some may transform into pre-cancerous lesions, ultimately progressing to colorectal cancer\cite{vazquez2017benchmark}. The timely detection and removal of these polyps through colonoscopy is thus of paramount importance in the prevention and management of CRC. Recognized as the gold standard, colonoscopy enables the identification and excision of polyps before they can advance to a more dangerous stage.

\begin{figure}[!ht]
\begin{center}
  \includegraphics[width=8.5cm,height=5.5cm]{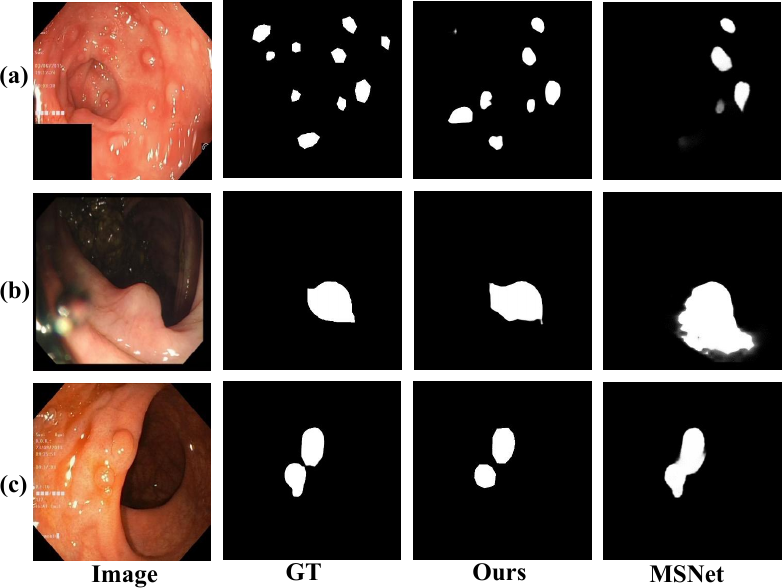}
\end{center}
   \caption{
Our proposed PSTNet model has been comprehensively evaluated and compared with MSNet\cite{zhao2021automatic} on a diverse set of challenging polyp images. These images include scenarios where the polyps are diminutive and easily overlooked (a), as well as situations where the segmentation boundaries are prone to errors due to blurred demarcations (b) and (c). The results of our experimental analyses demonstrate that PSTNet outperforms in terms of polyp localization capabilities and achieves higher segmentation accuracy.
}
\label{fig:cases}

\end{figure}

However, the accurate identification and segmentation of polyps during colonoscopy remain challenging due to the diverse sizes, shapes, and textures of polyps, coupled with their low contrast against surrounding tissues, leading to a camouflaging effect\cite{bernal2015wm,fan2021concealed}. This can result in missed or misdiagnosed polyps, severely compromising patient health outcomes.

In the early stages, polyp segmentation methods heavily relied on handcrafted features\cite{bernal2012towards}. However, these methods suffered from limited accuracy and generalization ability due to the restricted expressive power of handcrafted features and the inherent similarity between polyps and surrounding tissues. To enhance the accuracy and efficiency of polyp segmentation, researchers have developed a variety of deep learning (DL) architectures, employing different techniques to address this complex task. Encoder-decoder-based models, such as U-Net\cite{ronneberger2015u}, UNet++\cite{zhou2018unet++}, and ResUNet++\cite{jha2019resunet++}, along with attention-based approaches, including PraNet\cite{fan2020pranet}, Polyp-PVT\cite{dong2021polyp}, and SegT\cite{segt}, have significantly advanced the field of polyp segmentation by effectively capturing relevant features.
When compared with conventional methods, deep learning techniques have demonstrated significant advancements in segmentation. However, two challenging issues still persist:
(1) The inherent low contrast between polyps and their surrounding tissue renders them adept at camouflaging themselves, as discussed in the study by Fan et al.\cite{fan2021concealed}. This characteristic poses significant challenges in accurately localizing polyps, as illustrated in Figure \ref{fig:cases}(a). Moreover, polyp images often exhibit unclear boundaries, both between adjacent polyps and in the transition from polyps to normal tissue\cite{fan2020pranet}. These unclear boundaries give rise to segmentation inaccuracies, as evident in Figure \ref{fig:cases}(b) and Figure \ref{fig:cases}(c). Addressing these challenges is crucial for improving the effectiveness and reliability of polyp segmentation methods.
(2) Existing methodologies for polyp segmentation predominantly rely on feature data sourced exclusively from the RGB domain. However, relying solely on RGB features may not provide sufficient discriminative information to accurately distinguish between polyps and surrounding tissue, especially in cases of low contrast and unclear boundaries. Furthermore, discrepancies in pixel positions of features across different scales during the fusion process can significantly influence the accuracy of segmentation results. To address these limitations, it is crucial to explore alternative feature sources that can provide complementary information and address issues related to feature alignment and scale consistency, thereby enhancing the efficacy of polyp segmentation.

To address the previously mentioned issues, we have proposed a novel polyp segmentation network, which addresses the challenges of low contrast and indistinct boundaries in polyp segmentation by incorporating frequency domain cues. Our main contributions are as follows:

\begin{itemize}
    \item We propose \textbf{Polyp Segmentation Network
with Shunted Transformer (PSTNet)}, a novel deep learning architecture tailored for accurate polyp segmentation in colonoscopy images. The architecture incorporates frequency domain information, enhancing the model's capability to distinguish polyps from surrounding tissues.

    \item We introduce the \textbf{Feature Supplementary Alignment Module (FSAM)}, which employs feature alignment techniques to mitigate noise and achieve precise delineation of polyp boundaries. FSAM leverages multi-scale subtraction to extract distinctions between features at individual scales, thereby improving feature quality and addressing the issue of unclear boundaries.
    
    \item  We establish the \textbf{Frequency Characteristic Attention Module (FCAM)}, which infuses frequency cues from low-level features into the feature representation. This fusion enriches the available feature information, enabling it to be combined with global features for accurate localization of polyp regions.
    
    \item We design the \textbf{Cross Perception localization Module (CPM)} to interconnect the features generated by PSTNet's constituent components, leading to precise segmentation outcomes. By integrating the enhanced features from FCAM and FSAM, CPM ensures the effective utilization of comprehensive feature information, allowing PSTNet to attain state-of-the-art performance.

    \item Through extensive experimentation on five challenging datasets, we demonstrate that PSTNet significantly outperforms most models, establishing a new benchmark for polyp segmentation. The superior performance validates the effectiveness of our proposed modules and network architecture in addressing key challenges, highlighting PSTNet's potential to improve the accuracy and reliability of polyp segmentation in clinical practice.
    
\end{itemize}

\section{Related work}

\subsection{Hand-Crafted Methods}
Traditional automated polyp detection systems primarily rely on manual feature extraction techniques. These techniques encompass various methods, such as shape-based\cite{iwahori2017automatic}, texture-based\cite{ameling2009texture}, valley-depth-based\cite{bernal2015wm}, and combined approaches\cite{fu2014feature}.
However, due to the strong intra-class variability and weak inter-class differences between polyp regions and highly similar regions, the representational ability of the extracted features by these handcrafted methods is rather limited.
Consequently, these methods carry substantial risks of missed detections and false positives, necessitating the exploration of alternative, more effective approaches to achieve accurate polyp detection.

\subsection{Deep Learning-Based Methods}
\subsubsection{CNN-Based Polyp Segmentation}

Deep learning has revolutionized medical image analysis\cite{zhao2021attention,gui2022design,gui2023soft,dong2016left,luo2016novel,xu2022instance,wang2022net,wang2021retinal}, particularly in the domain of polyp detection. The MICCAI 2015 Colonoscopy Video Challenge showcased the superiority of CNN-based methods over traditional hand-crafted approaches, with the top-performing CNN achieving higher precision and recall values\cite{bernal2017comparative}.

Encoder-decoder architectures, such as U-Net\cite{ronneberger2015u}, UNet++\cite{zhou2018unet++}, and ResUNet++\cite{jha2019resunet++}, have gained popularity in medical image analysis due to their exceptional performance. DDA-Net\cite{tomar2021ddanet}, a two-decoder attention network built upon ResUNet++, demonstrated its effectiveness in polyp segmentation on the Kvasir-SEG dataset, achieving high dice coefficient and mIoU values.
Various advancements have been made in CNN-based architectures for polyp image segmentation, including the use of parallel LSTM with DeepLab-v3\cite{xiao2018semantic}, multi-task segmentation models\cite{murugesan2019psi}, reverse attention modules\cite{fan2020pranet}, and dual-tree wavelet pool CNNs\cite{banik2020polyp}. ADSNet\cite{adsnet} adapted to diverse semantic nuances in ambiguous polyp segmentation areas using a complementary trilateral decoder and continuous attention module.
Further developments include real-time polyp segmentation with ColonSNet\cite{jha2021real}, the application of generative adversarial networks\cite{ahmed2020generative}, and the use of contextual pixel relations and attention mechanisms in DCRNet\cite{yin2022duplex}. MSNet\cite{zhao2021automatic} and EUNet\cite{patel2021enhanced} addressed polyp size variability and image noise, while PolySeg Plus\cite{PolySegPP} utilized active learning to ameliorate data limitations and false positives. DUCK-Net\cite{Duck-net} employed residual downsampling and custom convolutional blocks to extract multi-resolution features.

Despite these advancements, CNN-based methods still face challenges in capturing long-range dependencies and efficiently processing high-resolution images, which are crucial for accurate polyp segmentation. While promising results have been achieved, there remains room for improvement in addressing the variability in polyp size, shape, and appearance, as well as dealing with image noise and artifacts.

\subsubsection{Transformer-Based Polyp Segmentation}

The transformer architecture, initially designed for machine translation, has been adapted for vision tasks, achieving remarkable performance\cite{xu2024mrftrans,zhang2024navid,xu2021dc,zhang2023task,wang2023treating,wang2023accurate}. The vision transformer (ViT)\cite{dosovitskiy2020image} partitions an image into patches, encodes them, and feeds them sequentially to the transformer encoder, performing image classification using a multi-layer perceptron. Compared to traditional CNNs, ViT models offer advantages such as handling larger input sizes, capturing global dependencies between pixels, and faster convergence with larger batch sizes.

ViT has also been applied to segmentation tasks\cite{Xu2023RSSFormerFS,xu2023scd,xu2023dual,Xu2024HCFNetHC,Xu2023SpectralPT}. To address intensive prediction tasks, pyramidal structures have been incorporated into transformers, as seen in models like PVT and shunted transformer, which utilize hierarchical transformers with multiple stages. Recently, the transformer architecture has been applied to polyp segmentation, with models like Polyp-PVT\cite{dong2021polyp} combining multiscale features from PVTv2 and a CNN-based decoder for accurate polyp segmentation. Other works like Duplex\cite{yin2022duplex}, TGANet\cite{tomar2022tganet}, PraNet\cite{fan2020pranet}, and SegT\cite{segt} also utilize attention for polyp segmentation tasks, exploring techniques like inverse attention, adversarial training, and edge guidance.

While transformer-based methods have shown promising results in polyp segmentation, they are still relatively new in this field and face challenges such as high computational costs, the need for large amounts of training data, and interpretability concerns. To address the key challenges of polyp segmentation, such as scale inconsistency, low contrast, and unclear boundaries, we propose the Polyp Segmentation Shunted Transformer Network (PSTNet). 
PSTNet introduces innovative modules to effectively align multi-scale features, integrate frequency information from polyp images, and improve polyp localization and segmentation accuracy.

\begin{figure*}[!ht]

\begin{center}
  \includegraphics[width=15cm,height=6cm]{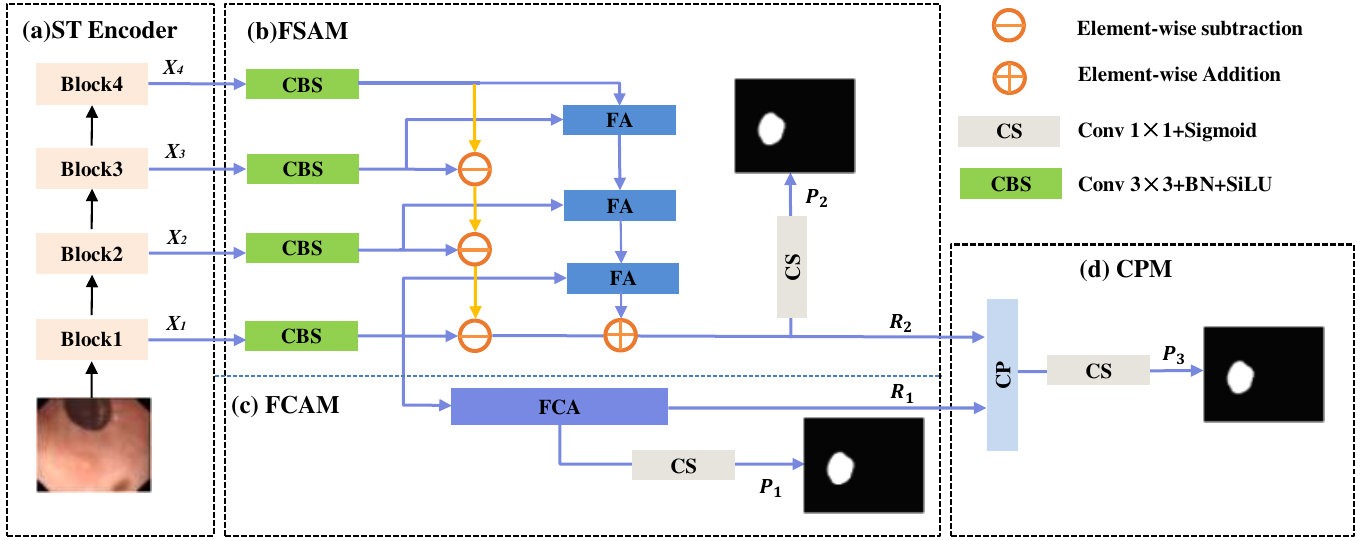}
\end{center}
   \caption{The framework of our PSTNet, which includes the shunted transformer (ST)\cite{ren2022shunted} (a) as an encoder network, (b) Feature Supplementary Alignment Module (FSAM) for fusing global semantic features,which contains three Feature Alignment (FA) units, (c) Frequency Characteristic Attention module (FCAM) for extracting low-level semantic features with frequency domain cues, and (d) Cross Perception localization Module (CPM) for linking frequency domain cues with global semantic features for the final output.}
\label{fig:ssanet}

\end{figure*}

\section{Proposed Method}
The PSTNet is a polyp segmentation architecture that utilizes a multi-scale feature fusion framework with a shunted transformer. The goal of this design is to enhance feature representation by aligning features and extracting more detailed information through the combination of frequency-domain cues. The process of the proposed method is shown in Algorithm~\ref{alg:proposed_network}. In the following sections, we discuss the methods used in the network in more detail.

\subsection{Overall Architecture}
\label{ssanet}
The proposed method (shown in Fig.~\ref{fig:ssanet}) comprises four primary modules: shunted transformer encoder (ST Encoder), feature supplementary alignment module (FSAM), frequency characteristic attention module (FCAM), and cross perception localization module (CPM).
The shunted transformer is utilized to extract feature maps at four different scales ($\mathbf{X}_{1}$, $\mathbf{X}_{2}$, $\mathbf{X}_{3}$, and $\mathbf{X}_{4}$) from the input image, spanning from low to high levels.
FCAM extracts shallow-level features from the low-level feature image $\mathbf{X}_{1}$, enabling the capture of polyps in various sizes and shapes through frequency domain analysis. The resulting intermediate output is denoted as $\mathbf{P}_{1}$.
Furthermore, the feature maps at the four different scales undergo convolutional operations to match the desired number of channels. Subsequently, FSAM is employed to progressively align and fuse features from all levels, yielding the intermediate result $\mathbf{P}_{2}$.
CPM integrates features from both FCAM and FSAM, enabling the efficient fusion of low-level semantics (containing frequency domain cues) with global semantics. This fusion yields the prediction $\mathbf{P}_{3}$ as the final output. Additionally, the sum of $\mathbf{P}_{1}$ and $\mathbf{P}{2}$ is computed, weighted, and added to $\mathbf{P}_{3}$ to generate the ultimate prediction.
In the training phase, we optimized the model using a primary loss function $\mathcal{L}_{3}$ as well as auxiliary loss functions $\mathcal{L}_{1}$ and $\mathcal{L}_{2}$. The primary loss was computed by comparing the final segmentation result $\mathbf{P}_{3}$ with the ground truth (GT), serving as the optimization target for achieving accurate polyp segmentation. Similarly, the auxiliary losses supervised the intermediate outputs, $\mathbf{P}_{1}$ and $\mathbf{P}_{2}$, generated by FCAM and FSAM, respectively.

\begin{algorithm}[H]
\caption{Proposed PSTNet for Polyp Segmentation}
\label{alg:proposed_network}
\begin{algorithmic}[1]
\REQUIRE Input image $I$
\STATE Extract feature maps $\{\mathbf{X}_{1}, \mathbf{X}_{2}, \mathbf{X}_{3}, \mathbf{X}_{4}\}$ from $I$ using ST Encoder
\STATE Obtain $\mathbf{P}_{1}$ from $\mathbf{X}_{1}$ using FCAM
\STATE Match channel dimensions of $\{\mathbf{X}_{1}, \mathbf{X}_{2}, \mathbf{X}_{3}, \mathbf{X}_{4}\}$ using convolutions
\STATE Obtain $\mathbf{P}_{2}$ by progressively aligning and fusing $\{\mathbf{X}_{1}, \mathbf{X}_{2}, \mathbf{X}_{3}, \mathbf{X}_{4}\}$ using FSAM
\STATE Integrate $\mathbf{P}_{1}$ and $\mathbf{P}_{2}$ using CPM to obtain $\mathbf{P}_{3}$
\STATE Compute weighted sum of $\mathbf{P}_{1}$ and $\mathbf{P}_{2}$ and add to $\mathbf{P}_{3}$ to obtain final prediction
\STATE Optimize model using primary loss $\mathcal{L}_{3}$ on $\mathbf{P}_{3}$ and auxiliary losses $\mathcal{L}_{1}$, $\mathcal{L}_{2}$ on $\mathbf{P}_{1}$, $\mathbf{P}_{2}$
\ENSURE Polyp segmentation prediction
\end{algorithmic}
\end{algorithm}

\subsection{Shunted Transformer Encoder for PSTNet}
\label{st}

In the context of polyp images, noise can be a significant issue due to various uncontrolled factors in the image acquisition process. To address this, we utilized a shunted transformer as the backbone network, leveraging its higher performance and better robustness to input interference compared to other methods like ViT. The shunted transformer allows for the extraction of more robust features from polyp images.
Polyps can exhibit diverse characteristics such as different sizes, shapes, and appearances. Therefore, it is essential to have an effective multi-scale representation for feature extraction. The shunted transformer, as described in the study by Ren et al.\cite{ren2022shunted}, employs a unique strategy called shunted self-attention (SSA). This approach enables the ViT model to incorporate attention at mixed scales within each attention layer. It effectively models objects of different scales simultaneously by assigning different attention heads in the same layer. This approach provides computational efficiency while preserving the ability to capture fine-grained details, which is crucial for detecting small polyps that might otherwise be overlooked.
To accommodate the polyp segmentation task, we generated four multi-scale feature maps (i.e., $\mathbf{X}_{1}$, $\mathbf{X}_{2}$, $\mathbf{X}_{3}$, and $\mathbf{X}_{4}$) at different stages of the shunted transformer and designed subsequent modules based on them.

\subsection{Frequency Characteristic Attention Module}
\label{sec:fcam}

While low-level RGB features provide detailed visual information about polyps, such as texture, color, and boundaries, polyps often remain concealed within normal tissues, posing a challenge for visual detection solely based on human perceptual capabilities\cite{zhong2022detecting}. To surpass the limitations of human biological vision, it is essential to incorporate additional cues beyond the RGB domain.
The FcaNet approach\cite{qin2021fcanet} operated in the frequency domain by extending the concept of global average pooling (GAP) to a 2D discrete cosine transform representation, enabling the effective utilization of additional frequency components for more comprehensive data analysis.
Therefore, we introduce frequency domain information as an additional cue to better distinguish polyps from the background and reduce the probability of false and missed detections.

\begin{figure*}[!ht]

\begin{center}
  \includegraphics[width=14cm,height=4cm]{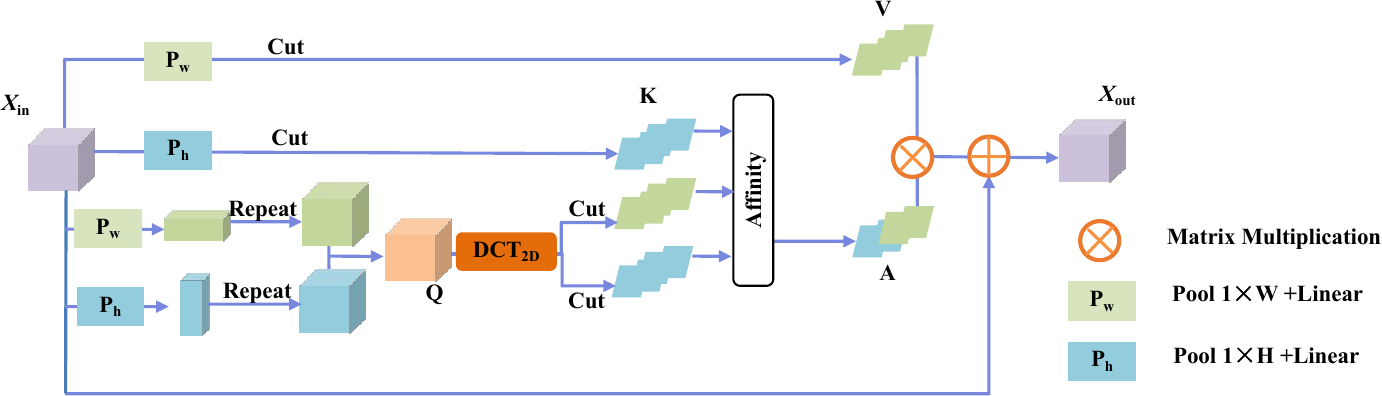}
\end{center}
   \caption{The details of the frequency characteristic attention module (FCAM). First, the input ${X}_{in}$ is cut, repeated and fused horizontally and vertically so that each spatial location obtains a feature response from a global context with the same horizontal and vertical coordinates. Secondly, we combined a 2D discrete cosine transform (${DCT}_{2D}$) to obtain spectral information, and finally, we used the resulting full attention affinity to re-weight each channel map.}

\label{fig:fcam}

\end{figure*}
We designed the frequency domain perceptual attention module, as shown in Figure \ref{fig:fcam}, to capture the details of polyps from the combination of both the RGB domain and frequency domain of the low-level feature $\mathbf{X}_{1}$, which can accurately locate the position of polyps.
Specifically, the FCAM includes the full attention enhancement operation (${Att}_{f}$) and the 2D discrete cosine transform (${DCT}_{2D}$), the ${Att}_{f}$ operation compositional a function can be defined as:
\begin{align}
\label{equ:attf}
{Att}_{f}(\mathbf{X})=\alpha \cdot ( A\left (\mathbf{Q},\mathbf{K} \right )  \cdot \mathbf{V})+\mathbf{X}
 \end{align}
where $\alpha$ is the scale parameter, $X \in \mathbb{R}^{C \times H \times W}$ is the input low-level feature, and $A\left ( \cdot  \right ) $ is the affinity operation, which is defined as follows:
\begin{align}
A_{i, j}=\frac{\exp \left(\mathbf{Q}_{i} \cdot \mathbf{K}_{j}\right)}{\sum_{i=1}^{C} \exp \left(\mathbf{Q}_{i} \cdot \mathbf{K}_{j}\right)}
 \end{align}
where $A_{i, j} \in A$ denotes the degree of correlation between the $i_{th}$ and $j_{th}$ channel at a specific spatial position.

In the 2D Discrete Cosine Transform (DCT), basis functions are defined as follows:
\begin{equation}
B_{h,w}^{i,j} = \cos\left(\frac{\pi h}{H}\left(i+\frac{1}{2}\right)\right)\cos\left(\frac{\pi w}{W}\left(j+\frac{1}{2}\right)\right)
\label{eq_basis}
\end{equation}
The 2D DCT computation is expressed as:
\begin{equation}
\begin{aligned}
& DCT_{2D} = \sum_{i=0}^{H-1} \sum_{j=0}^{W-1} \mathbf{x}_{i,j}^{2d} B_{h,w}^{i,j}\\
        s.t. \;\; h \in \{0,1, & \cdots,H-1\}, w \in \{0,1,\cdots,W-1\},
\end{aligned}
\label{eq_2ddct}
\end{equation}
Here, $DCT_{2D}$ is the outcome of the 2D DCT computation, and $\mathbf{x}^{2d} \in \mathbb{R}^{H \times W}$ is the input signal.

We further process $\mathbf{X}$ to obtain $\mathbf{Q}_{P}$:
\begin{align}
\mathbf{Q}_{P}=[\operatorname{Re}_{h}\left(\operatorname{CV}_{w}\right), \operatorname{Re}_{w}\left(\operatorname{CV}_{h}\right)]
\end{align}
Here, ${CV}_{w}$ and ${CV}_{h}$ represent pooling along the width ($W$) and height ($H$) dimensions, respectively, followed by a linear layer. ${Re}_{h}$ and ${Re}_{w}$ indicate replication along the height and width directions, and $[\cdot]$ signifies the concatenation of tensors.
Next, we split $\mathbf{Q}_{P}$ into multiple parts along the channel dimension, denoted as $[\mathbf{Q}_{P}^0,\mathbf{Q}_{P}^1, \cdots, \mathbf{Q}_{P}^{n-1}]$, where $\mathbf{Q}_{P}^i \in \mathbb{R}^{C' \times H \times W}$, $i \in {0,1, \cdots, n-1}$,$C' = {C}\times {n}^{-1}$ , and $C$ is divisible by $n$. Each part corresponds to a 2D DCT frequency component, resulting in:
\begin{equation}
\begin{aligned}
\mathbf{Freq}^i & = DCT_{2D}^{u_i,v_i}(\mathbf{Q}_{P}^i) \
& s.t. ;; i \in {0,1,\cdots,n-1}
\end{aligned}
\label{eq_2ddct_network}
\end{equation}
where $[u_i,v_i]$ are the frequency component 2D indices corresponding to $X^i$, and $\mathbf{Freq}^i \in \mathbb{R}^{C'}$ is the compressed vector.

Finally, we concatenate these frequency components to obtain the multi-spectral vector:
\begin{equation}
\begin{aligned}
\mathbf{Q} = \text{Cat}([ \mathbf{Freq}^0, \mathbf{Freq}^1,\cdots, \mathbf{Freq}^{n-1}])
\end{aligned}
\label{eq_cat}
\end{equation}
Here, $\mathbf{Q}$ is the resulting multi-spectral vector. In these ways, the FCAM introduces multispectral channel information while modeling attention spatially and in terms of channels. 

\subsection{Feature Supplementary Alignment Module}
\label{sec:fsam}
Many contemporary polyp segmentation networks face a critical challenge arising from the mismatch problem induced by frequent downsampling operations and the unprocessed integration of contextual information during feature aggregation. The noise generated by pixel shifts during feature fusion across various scales exacerbates this issue. To address this concern, we introduce a novel solution: a feature alignment method specifically designed for cascaded feature fusion.

As shown in Figure \ref{fig:ssanet} (b), the inputs are $\mathbf{X}_{1}$, $\mathbf{X}_{2}$, $\mathbf{X}_{3}$, and $\mathbf{X}_{4}$, representing four scales of features. To ensure the robustness of the features, the neighboring features are cascaded and fused using the multi-scale subtraction unit (SU), while the FA unit is employed for feature alignment of the neighboring features. The resulting cascade fusions are then combined using element-wise addition to obtain $\mathbf{P}_{2}$. In our implementation, $f(\cdot)$ is specified as a convolutional unit, which comprises a $3 \times 3$ convolutional layer with padding set to 1, along with batch normalization and the SiLU activation function (CBS). The detailed process of the two cascaded fusions is as follows.

\begin{figure}[!ht]
\begin{center}
  \includegraphics[width=7cm,height=6cm]{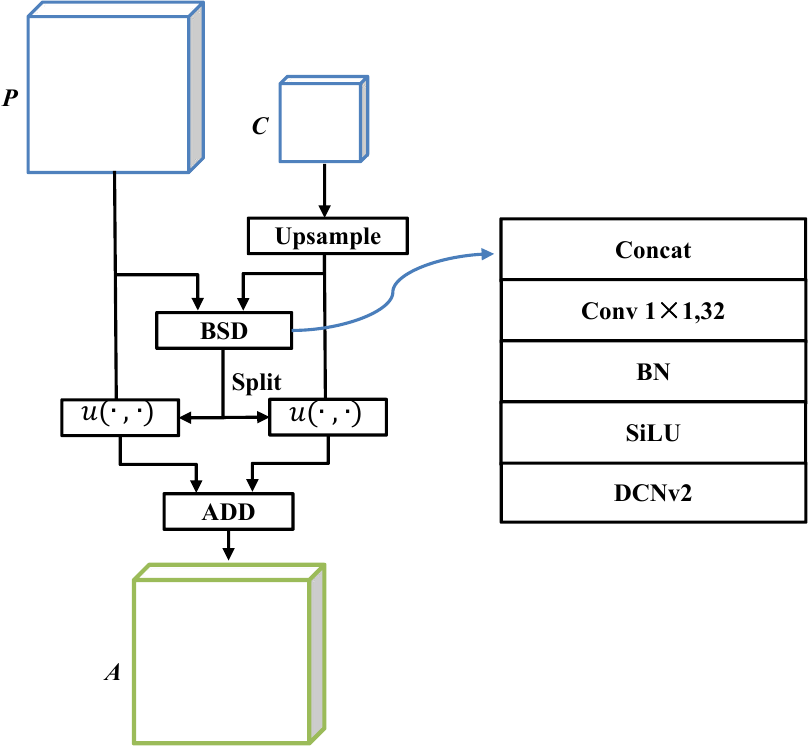}
\end{center}
   \caption{The details of the Feature Alignment units. First, the high-level features $\mathbf{C}$ are upsampled and connected to the neighbouring low-level features $\mathbf{P}$. The two predicted biased feature maps are then obtained by deformable convolution in BSD, and the features are aligned separately for both scales, followed by a summation operation.}
\label{fig:fa}

\end{figure}

\subsubsection{Supplementary Fusion}
The SU captures complementary information from adjacent feature layers and emphasizes their differences, providing the decoder with differential feature information. This can be expressed as:

\begin{align}
    SU=\left|f(\mathbf{X}_{i}) \ominus f(\mathbf{X}_{i-1})\right|
\end{align}
where $\ominus$ denote element-wise subtraction,$\left|\cdot \right|$ means calculate the absolute value.
We upsample $\mathbf{X}_{4}$ to the same size as $\mathbf{X}_{3}$, and then pass $f(\cdot )$ , the results are noted as ${\mathbf{X}_{4}^{\prime}}$, ${\mathbf{X}_{3}^{\prime}}$. Then using the SU unit to fuse ${\mathbf{X}_{4}^{\prime}}$ and  ${\mathbf{X}_{3}^{\prime}}$ to obtain $S_{1}$ can be expressed as
    $S_{1}=\left|f(\mathbf{X}_{4}^{\prime}) \ominus f(\mathbf{X}_{3}^{\prime})\right|$. 
Then, $\mathbf{S}_{1}$ is upsampled to the same size as $\mathbf{X}_{2}$, which is recorded as $S_{1}^{\prime}$, $\mathbf{X}_{2}$ is passed through $f(\cdot )$, the result is recorded as ${\mathbf{X}_{2}^{\prime}}$, and use the SU to fuse $\mathbf{S}_{1}^{\prime}$ and ${\mathbf{X}_{2}^{\prime}}$, the result is recorded as $\mathbf{S}_{1}$, which can be expressed as
    $\mathbf{S}_{2}=\left|f(\mathbf{S}_{1}^{\prime}) \ominus f(\mathbf{X}_{2}^{\prime})\right|$. 
Similarly, the final supplemental fusion result $S_{3}$ can be obtained, which can be expressed as 
    $\mathbf{S}_{3}=\left|f(\mathbf{S}_{2}^{\prime}) \ominus f(\mathbf{X}_{1}^{\prime})\right|$.

\subsubsection{Feature Align}

In the multi-scale feature fusion process, a natural spatial offset exists between the pixel positions of the upper feature $\mathbf{F}_{i}$ and its lower feature $\mathbf{F}_{i-1}$. This offset cannot be eliminated either through concatenation or elemental operations\cite{huang2021alignseg}.
In Figure \ref{fig:fa}, $\mathbf{C}_{i}$ is initially upsampled to obtain $\tilde{\mathbf{C}}_{i}$, which is then concatenated with $P$. These concatenated features are subsequently processed through DCNv2\cite{zhu2019deformable} with a kernel size of $3\times 3$, resulting in the generation of two offset maps, namely $\Delta_{C}$ and $\Delta_{P}$.

These offset maps play a pivotal role in the calibration of the low-resolution features, $\mathbf{C}_{i}$, and the high-resolution features, $\mathbf{P}$\cite{huang2021fapn}, respectively. Once the offset maps are obtained, our feature alignment aggregation can be difined as follows:

\begin{align}
    {A}_{i}={u}\left(\text {Upsample}\left(\tilde{\mathbf{C}}_{i}\right), \boldsymbol{\Delta}_{C}\right)+{u}\left(\tilde{\mathbf{P}}, \boldsymbol{\Delta}_{A}\right)
\end{align}
where upsample denotes a bilateral interpolation function, while ${u}(\cdot ,\cdot)$ represents the alignment function.  It can be assumed that the spatial coordinates of each position on the feature map $F$ to be aligned are given as ${(1,1),(1,2), \ldots,(H, W)}$, with the offset map represented by $\Delta \in {R}^{2 \times H \times W}$. The $U_{hw}$ is proposed in SFSegNet\cite{jiang2019sfsegnet}, which is the output of the alignment function ${u}\left(F, \boldsymbol{\Delta}\right)$ and alignment function is defined as follows:

\begin{equation}
\begin{array}{c} 
{U}_{h w}=\sum_{h^{\prime}=1}^{H} \sum_{w^{\prime}=1}^{W} {F}_{h^{\prime} w^{\prime}} \\
\cdot \max \left(0,1-\left|h+\Delta_{1 h w}-h^{\prime}\right|\right) \\
\cdot \max \left(0,1-\left|w+\Delta_{2 h w}-w^{\prime}\right|\right)
\end{array}
\end{equation}

which samples feature on position $\left(h+\Delta_{1 h w}, w+\Delta_{2 h w}\right)$ of $F$, using the bilinear interpolation kernel, where $\Delta_{1 h w}$, $\Delta_{2 h w}$ indicate the learned 2D transformation offsets for position $(h, w)$. 

\subsection{Cross Perception localization Module}
\label{sec:cpm}
\begin{figure}[htb]
\begin{center}
  \includegraphics[width=7cm,height=3cm]{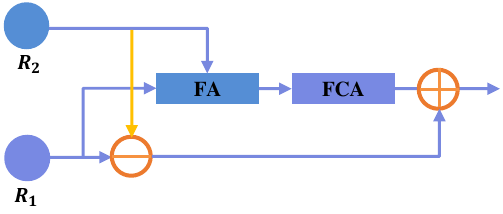}
\end{center}
   \caption{The details of the CP unit. The low-level features $R_{1}$ containing frequency domain information are aligned with the global features $R_{2}$ via the FA unit, then enhanced via the FCA unit, and finally added with the result of the subtraction of $|\mathbf{R}_{1}-\mathbf{R}_{2}|$.}
\label{fig:cp}

\end{figure}

To combine the spectral cues acquired from FCAM with the semantic information derived from FSAM, we introduced a cross-perception localization module, as illustrated in Figure \ref{fig:ssanet}. This module takes two input feature maps, $\mathbf{R}{2}$ containing global semantic information and $\mathbf{R}{1}$ with frequency domain information and rich details. These input feature maps are processed by the Cross-Perception (CP) unit, depicted in Figure \ref{fig:cp}.
The CP unit performs an element-wise subtraction operation on the features from the FCAM ($\mathbf{R}^{\varsigma}_1$) and FSAM ($\mathbf{R}^{\varsigma}_2$) branches, resulting in a feature map denoted as $\mathbf{R}^{\varsigma}$. Subsequently, the Feature Alignment (FA) unit (which described in Section \ref{sec:fsam}) aligns these features. The aligned feature map is then passed to the Frequency Characteristic Attention (FCA) unit, with the 2D DCT operation excluded. Finally, the feature map $\mathbf{R}^{\varsigma}$ is added to the output of the FCA unit to produce the resulting feature map $\mathbf{Z}$.
The entire process can be summarized as equation:
\begin{align}
    \mathbf{Z} = Att_{f/DCT}(f_{a}(\mathbf{R}_{1},\mathbf{R}_{2}))+\mathbf{R}^{\varsigma }  
\end{align}
where $f_{a}$ is the operation to perform feature alignment and $Att_{f/DCT}$ is the frequency domain aware attention operation to remove the 2D DCT transform.
$\mathbf{Z}$ by sigmoid to obtain the final segmentation result $\mathbf{P}_{3}$.

\subsection{Loss Function}
In medical image segmentation, class imbalance is a common challenge. The number of background pixels vastly exceeds the number of pixels belonging to the target objects. This imbalance can hinder the model's ability to learn effective representations, particularly for object boundaries. To address this issue, we adopted a combination loss function. Given the Dice loss's good performance in handling class imbalance \cite{sudre2017generalised}, we chose it as the base loss function. However, directly employing the Dice loss may lead to instability during training. To mitigate this issue, we introduced the Focal loss\cite{lin2017focal}, which concentrates the training process on pixels that are difficult to classify correctly, further enhancing the segmentation quality of object boundaries. Additionally, we incorporated the weighted Binary Cross-Entropy (wBCE) loss to further improve the segmentation accuracy, especially at object boundaries, where misclassification errors have a greater impact. We specifically designed this combination loss function to optimize model performance and address the complexities associated with challenging pixels and small object segmentation.
Given the ground-truth $G$ and the prediction $\mathbf{P}_{i},i \in\{1,2,3\}$, the
total loss function $\mathcal{L}_{\text {total}}$ is given by:
\begin{align}
    \mathcal{L}_{\text {total}}=\gamma \cdot \mathcal{L}_{\text 1}\left(\mathbf{Y}, {\mathbf{P}}_{1}\right)+\lambda\cdot\mathcal{L}_{\text 2} \left(\mathbf{Y}, {\mathbf{P}}_{2}\right)+\mathcal{L}_{\text 3}\left(\mathbf{Y}, {\mathbf{P}}_{3}\right)
\end{align}
where the parameter was set experimentally to $\gamma = 0.1$ ,$\lambda = 1$, each loss term is calculated by $L^{\psi }$, which is defined as:
\begin{align}
    {\mathcal{L}^{\psi}}=\mathcal{L}_{\text {bce}}^{w}+\mathcal{L}_{\text {dice }}+\mathcal{L}_{\text {focal }}
\end{align}
Here, $\mathcal{L}_{bce}^{w}$ represents a weighted binary cross-entropy (BCE) loss function. In contrast to the standard BCE loss, which treats all pixels equally, $\mathcal{L}_{bce}^{w}$ takes into account the importance of each pixel by assigning higher weights to challenging pixels. The weighting scheme enables the network to prioritize difficult regions, leading to an overall improvement in performance. Additionally, $\mathcal{L}_{\text{dice}}$ refers to the dice loss, while $\mathcal{L}_{\text{focal}}$ is the focal loss\cite{lin2017focal}. The dice loss effectively learns the class distribution, mitigating imbalanced voxel problems. Conversely, the focal loss compels the model to improve its learning for poorly classified pixels\cite{zhu2019anatomynet}.
 
\section{Experiment}
\label{sec:experiment}
\subsection{Datasets and Compared SOTA Methods}  
\subsubsection{Datasets}
Our approach underwent evaluation using five challenging colonoscopic polyp image datasets: Kvasir-SEG\cite{jha2020kvasir}, CVC-Clinic\cite{bernal2015wm}, ETIS\cite{silva2014toward}, CVC-ColonDB\cite{tajbakhsh2015automated}, and EndoScene\cite{vazquez2017benchmark} datasets. 
The EndoScene dataset is a combination of CVC-612 and CVC-300. In our experiments, we solely utilized theCVC-300 test set, as a portion of the CVC-612 dataset might have been employed for training purposes.

\begin{table*}[!ht]

\centering
\caption{Statistical comparison of the learning ability of CVC-ClinicDB and Kvasir SEG. ↑ and ↓ denote respectively that the larger and smaller scores are better.} The Best and second best scores are shown in \textcolor{red}{red} and \textcolor{blue}{blue} respectively.
\label{tab:compare-2}
\renewcommand{\arraystretch}{1.1}
\begin{tabular}{c|c|c|c|c|c|c|c|c} 
\hline
         & Model                 & $mDic(\%) \uparrow$                     & $mIoU(\%) \uparrow$            & $F_\beta^\omega(\%) \uparrow$  & $S_\alpha(\%) \uparrow$        & $m E_\epsilon(\%) \uparrow$    & $\max E_\epsilon(\%) \uparrow$ & $MAE(\%) \downarrow$            \\ 
\hline
         & U-Net++ (TMI'19)      & 82.1                                    & 74.3                           & 80.8                           & 86.2                           & 88.6                           & 90.9                           & 4.8                             \\
         & PraNet (MICCAI'20)    & 89.8                                    & 84.0                           & 88.5                           & 91.5                           & 94.4                           & 94.8                           & 3.0                             \\
Kvasir   & DCRNet (arXiv'21)     & 88.6                                    & 82.5                           & 86.8                           & 91.1                           & 93.3                           & 94.1                           & 3.5                             \\
         & SANet (MICCAI'21)     & 90.4                                    & 84.7                           & 89.2                           & 91.5                           & 94.9                           & 95.3                           & 2.8                             \\
         & MSNet (MICCAI'21)     & {90.7}                                    & {86.2}                           & {89.3}                           & \textcolor{blue}{92.2}                           & \textcolor{blue}{95.2}                           & \textcolor{blue}{94.4}                           & {2.8}                             \\
          & ADSNet (BMVC'24)  & {92.0}                                    & {87.1}                           & \textcolor{blue}{91.6}                             & ——                             & ——                             & ——                             & {2.3}                              \\ 
           & SegT (arXiv'23)  & \textcolor{blue}{92.7}                                    & \textcolor{blue}{88.0}                           & ——                             & ——                             & ——                             & ——                             & \textcolor{blue}{2.3}                              \\ 
\cline{2-9}
           & \textbf{PSTNet(Ours)} & \textcolor{red}{\textbf{93.5}}          & \textcolor{red}{\textbf{89.5}} & \textcolor{red}{\textbf{92.9}} & \textcolor{red}{\textbf{93.7}} & \textcolor{red}{\textbf{96.7}} & \textcolor{red}{\textbf{97.3}} & \textcolor{red}{\textbf{1.7}}   \\ 
\hline
         & U-Net++ (TMI'19)      & 79.4                                    & 72.9                           & 78.5                           & 87.3                           & 89.1                           & 93.1                           & 2.2                             \\
         & PraNet (MICCAI'20)    & 89.9                                    & 84.9                           & 89.6                           & 93.6                           & 96.3                           & 97.9                           & 0.9                             \\
CVC-ClinicDB & DCRNet (arXiv'21)     & 89.6                                    & 84.4                           & 89.0                           & 93.3                           & 96.4                           & 97.8                           & 1.0                             \\
         & SANet (MICCAI'21)     & 91.6                                    & 85.9                           & 90.9                           & 93.9                           & 97.1                           & 97.6                           & 1.2                             \\
         & MSNet (MICCAI'21)     & {92.1}                                    & 87.9                          & {91.4}                           & \textcolor{blue}{94.1}                           & \textcolor{blue}{97.6}                           & \textcolor{blue}{97.2}                           & \textcolor{blue}{0.8}                             \\
          & ADSNet (BMVC'24)  & 93.8                                    & 89.0                           & 94.0                             & ——                             & ——                             & ——                             & 0.6                             \\ 
           & SegT (arXiv'23)  & \textcolor{blue}{94.0}                                    & \textcolor{blue}{89.7}                          & ——                             & ——                             & ——                             & ——                             & \textcolor{red}{0.6}                             \\ 
\cline{2-9}
         & \textbf{PSTNet(Ours)} & \textcolor{red}{\textbf{94.5}}          & \textcolor{red}{\textbf{90.1}} & \textcolor{red}{\textbf{94.5}} & \textcolor{red}{\textbf{95.3}} & \textcolor{red}{\textbf{98.7}} & \textcolor{red}{\textbf{99.0}} & \textcolor{blue}{\textbf{0.7}}  \\
\hline
\end{tabular}

\end{table*}

\subsubsection{Compared Methods}
We compared the performance of recent state-of-the-art (SOTA) models for polyp image segmentation on seven different metrics, including UNet++\cite{zhou2018unet++}, PraNet\cite{fan2020pranet}, DCRNet\cite{yin2022duplex}, SANet\cite{wei2021shallow}, MSNet\cite{zhao2021automatic}, ADSNet\cite{adsnet}, and SegT\cite{segt}. For a fair comparison, we used their open-source code and default settings to evaluate the models on the same training and test sets, and generated the result maps.

\subsection{Experimental Setting and Evaluation Metrics}
\subsubsection{Experimental Setting}
\label{setting}

During the training phase, we incorporated a multi-scale method\cite{ren2022shunted} to handle the size variations among individual polyp images. The AdamW optimizer, a common choice for transformer networks, was employed to optimize the model parameters. Across all experiments, the model was trained for approximately 135 epochs, with a fixed learning rate of $1e-4$ and a decay rate of 0.1 every 45 epochs. For more specific parameter settings, please refer to Table~\ref{tab:setting}. To ensure the stability and effectiveness of the results, we conducted five independent training and testing runs.

\begin{table}[!ht]
\centering
\caption{The setting of parameters during training.}
\label{tab:setting}
\renewcommand{\arraystretch}{1.4}
\begin{tabular}{c|c|c} 
\hline
\textbf{Optimizer}          & \textbf{Decay rate}   & \textbf{Epochs}  \\
AdamW                       & 0.1                   & 135              \\ 
\hline
\textbf{Learning rate (lr)} & \textbf{Weight decay} & \textbf{Clip}    \\
1e-4                        & 1e-4                  & 0.5              \\
\hline
\textbf{Input size} & \textbf{Decay epoch} & \textbf{Batch size}    \\
352×352                        & 45                  & 20              \\
\hline
\end{tabular}

\end{table}

Our model was implemented using the PyTorch framework, and all training and testing experiments were conducted on a server equipped with four NVIDIA GeForce RTX 3090 GPUs. During the evaluation phase, images were scaled to a fixed size of $352 \times 352$ without applying any post-processing optimization techniques.

\subsubsection{Evaluation Metrics}

We employed six standard evaluation metrics commonly used in image segmentation tasks: mDic, IoU (Intersection over Union), S-measures ($S_{\alpha}$), weighted F-measure $F_{\beta}^{\omega}$, E-measure ($E_{\epsilon}$), and mean absolute error (MAE) to comprehensively evaluate the model performance. For a rigorous and fair comparison, all models were trained, validated, and tested using identical data splits. Additionally, we utilized pre-trained backbones from the ImageNet dataset and incorporated the authors' provided source code when available. This consistent methodology in metric selection and model evaluation enhances the reliability and accuracy of our comparative analysis.

\subsection{Performance Comparison}
\label{fig:compare_sota}
\subsubsection{Quantitative Results}
\begin{table*}[!ht]
\centering
\caption{Statistical comparison of the generalization ability of CVC-ColonDB, ETIS, and CVC-300. ↑ and ↓ denote respectively that the larger and smaller scores are better.} The best and second best scores are shown in \textcolor{red}{red} and \textcolor{blue}{blue} respectively.
\label{tab:com-unseen}
\renewcommand{\arraystretch}{1.1}
\begin{tabular}{c|c|c|c|c|c|c|c|c} 
\hline
\multicolumn{1}{c|}{} & \multicolumn{1}{c|}{Model} & $mDic(\%) \uparrow$            & $mIoU(\%) \uparrow$            & $F_\beta^\omega(\%) \uparrow$  & $S_\alpha(\%) \uparrow$        & $m E_\epsilon(\%) \uparrow$    & $\max E_\epsilon(\%) \uparrow$ & $MAE(\%) \downarrow$           \\ 
\hline
                     
                      & U-Net++ (TMI'19)           & 48.3                           & 41.0                           & 46.7                           & 69.1                           & 68.0                           & 76.0                           & 6.4                            \\
                     
                      & PraNet (MICCAI'20)         & 71.2                           & 64.0                           & 69.9                           & 82.0                           & 84.7                           & 87.2                           & 4.3                            \\
CVC-ColonDB           & DCRNet (arXiv'21)          & 70.4            & 63.1          & 68.4            & 82.1        & 84.0                       & 84.8                           & 5.2                            \\
                      & SANet (MICCAI'21)          & 75.3                           & 67.0                           & 72.6                           &  \textcolor{blue}{83.7}                           & 86.9                           & 87.8                           & 4.3                            \\
                      & MSNet (MICCAI'21)          & {75.5}                           & {67.8}                           & {73.7}                           & 83.6                         & \textcolor{blue}{87.0}                           & \textcolor{blue}{88.3}                           & {4.1}                            \\
                       & ADSNet (BMVC'24)  & \textcolor{blue}{81.5}                                    & {73.0}                           & \textcolor{blue}{86.0}                             & ——                             & ——                             & ——                             & {2.9}                              \\ 
                        & SegT (arXiv'23)  & {81.4}                                    & \textcolor{blue}{73.2}                           & --                             & ——                             & ——                             & ——                             & \textcolor{blue}{2.6}                              \\ 
                                         
\cline{2-9}
                      & \textbf{PSTNet(Ours)}      & \textcolor{red}{\textbf{82.7}} & \textcolor{red}{\textbf{74.8}} & \textcolor{red}{\textbf{81.3}} & \textcolor{red}{\textbf{87.7}} & \textcolor{red}{\textbf{92.5}} & \textcolor{red}{\textbf{92.8}} & \textcolor{red}{\textbf{2.5}}  \\ 
\hline
                     
                      & U-Net++ (TMI'19)           & 70.7                           & 62.4                           & 68.7                           & 83.9                           & 83.4                           & 89.8                           & 1.8                            \\
                  
                      & PraNet (MICCAI'20)         & 87.1                           & 79.7                           & 84.3                           & 92.5                           & 95.0                           & 97.2                           & 1.0                            \\
CVC-300               & DCRNet (arXiv'21)          & 85.6                           & 78.8                           & 83.0                           & 92.1                           & 94.3                           & 96.0                           & 1.0                            \\
                      & SANet (MICCAI'21)          & {88.8}                           & {81.5}                           & \textcolor{blue}{85.9}                           & \textcolor{blue}{92.8}                           & \textcolor{blue}{96.2}                           & \textcolor{blue}{97.2}                           &  \textcolor{blue}{0.8}                            \\
                      & MSNet (MICCAI'21)          & 86.9                          & 80.7                           & 84.9                           & 92.5                           & 95.8                          & 94.3                           & 1.0                            \\
                       & ADSNet (BMVC'24)  & ——                                    & ——                           & ——                             & ——                             & ——                             & ——                             & ——                              \\ 
                       & SegT (arXiv'23)  & \textcolor{blue}{89.5}                                    & \textcolor{blue}{82.8}                           & ——                             & ——                             & ——                             & ——                             & \textcolor{blue}{0.8}                              \\ 
                    
\cline{2-9}
                      & \textbf{PSTNet(Ours)}      & \textcolor{red}{\textbf{91.0}} & \textcolor{red}{\textbf{84.7}} & \textcolor{red}{\textbf{89.5}} & \textcolor{red}{\textbf{94.1}} & \textcolor{red}{\textbf{97.6}} & \textcolor{red}{\textbf{98.3}} & \textcolor{red}{\textbf{0.5}}  \\ 
\hline
                    
                      & U-Net++ (TMI'19)           & 40.1                           & 34.4                           & 39.0                           & 68.3                           & 62.9                           & 77.6                           & 3.5                            \\
                  
                      & PraNet (MICCAI'20)         & 62.8                           & 56.7                           & 60.0                           & 79.4                           & 80.8                           & 84.1                           & 6.7                            \\
ETIS                  & DCRNet (arXiv'21)          & 55.6                           & 49.6                           & 50.6                           & 73.6                           & 74.2                           & 77.3                           & 9.6                            \\
                      & SANet (MICCAI'21)          & \textcolor{blue}{75.0}                           &65.4                           & \textcolor{blue}{68.5}                           & \textcolor{blue}{84.9}                           & \textcolor{blue}{88.1}                           & \textcolor{blue}{89.7}                           & \textcolor{blue}{1.5}          \\
                      & MSNet (MICCAI'21)          & 71.9                           &  \textcolor{blue}{66.4}                           & 67.8                           & 84.0                           & 87.5                           & 83.0                           & 2.0                            \\
                      & ADSNet (BMVC'24)  & \textcolor{blue}{79.8}                                    & \textcolor{blue}{71.5}                           & \textcolor{blue}{79.2}                             & ——                             & ——                             & ——                             & \textcolor{blue}{1.2}                              \\ 
                       & SegT (arXiv'23)  & \textcolor{blue}{81.0}                                    & \textcolor{blue}{73.2}                           & ——                             & ——                             & ——                             & ——                             & \textcolor{blue}{1.3}                              \\ 
                     
\cline{2-9}
                      & \textbf{PSTNet(Ours)}      & \textcolor{red}{\textbf{80.0}} & \textcolor{red}{\textbf{72.6}} & \textcolor{red}{\textbf{76.1}} & \textcolor{red}{\textbf{87.5}} & \textcolor{red}{\textbf{90.1}} & \textcolor{red}{\textbf{91.3}} & \textcolor{red}{\textbf{1.3}}  \\
\hline
\end{tabular}

\end{table*}

\textbf{Learning Capability: } We selected two datasets, CVC-ClinicDB and Kvasir-SEG, as benchmarks. CVC-ClinicDB contains 612 images extracted from 31 colonoscopy videos, while Kvasir-SEG consists of 1000 polyp images collected from the polyp category of the Kvasir dataset. Following the practice of PraNet, we used 548 images from CVC-ClinicDB and 900 images from Kvasir-SEG as the training set, with the remaining 64 and 100 images serving as the test set for each dataset, respectively. As shown in Table \ref{tab:compare-2}, our model achieved the best results on both datasets, surpassing the state-of-the-art methods. On Kvasir-SEG, our model outperformed ADSNet and SegT by 1.5\% and 0.8\% in terms of mDice, and 2.4\% and 1.5\% in terms of mIoU, respectively. Similarly, on CVC-ClinicDB, our model matched the performance of TGANet in mDice and outperformed ADSNet and SegT by 0.7\% and 0.5\%, respectively, while achieving the highest mIoU of 90.1\%, demonstrating its strong learning capability. This can be attributed to our proposed multi-scale feature fusion framework, which enhances the model's ability to capture comprehensive feature information by combining information from both the RGB and frequency domains. Furthermore, the FCAM incorporates frequency cues from low-level features into global features, enriching the available feature information and aiding in the precise localization of polyp regions. 
These innovations enable our model to better learn the feature representations of polyps, outperforming other methods.

\textbf{Generalization Capabilities:} To comprehensively evaluate the generalization capability of the model, we employed three datasets that the model had never encountered before: CVC-ColonDB (380 images), CVC-300 (60 images), and ETIS (196 images). It is worth noting that this differs from the validation method used for the ClinicDB and Kvasir-SEG datasets, as the model had no exposure to these datasets during the training process. As shown in Table \ref{tab:com-unseen}, our proposed method still achieved the best results. On the CVC-ColonDB dataset, our model outperformed ADSNet and SegT by 1.2\% and 1.3\% in terms of mDice, and 1.8\% and 1.6\% in terms of mIoU, respectively. Similarly, on the CVC-300 dataset, our model surpassed SANet and SegT by 2.2\% and 1.5\% in mDice, and 3.2\% and 1.9\% in mIoU, respectively. Particularly on the most challenging ETIS dataset, where most images contain small polyps, our method outperformed the state-of-the-art ADSNet and SegT by 0.2\% and 1.0\% in terms of mDice, and 1.1\% and 0.6\% in terms of mIoU, respectively, demonstrating its exceptional generalization capability.
This can be primarily attributed to our proposed FSAM). FSAM employs feature alignment techniques to mitigate noise and utilizes multi-scale subtraction to extract the differences between features at various scales, enabling more precise delineation of polyp boundaries. This effectively addresses the issue of indistinct polyp boundaries, allowing our model to better adapt to unseen datasets. 
Moreover, the design of the Cross-Perception Module (CPM) also plays a crucial role, as it connects the features generated by all components of PSTNet and integrates the enhanced features from FCAM and FSAM, ensuring the effective utilization of comprehensive feature information and achieving superior generalization performance compared to other methods.

\begin{figure*}[!htp]

\begin{center}
  \includegraphics[width=16cm,height=13cm]{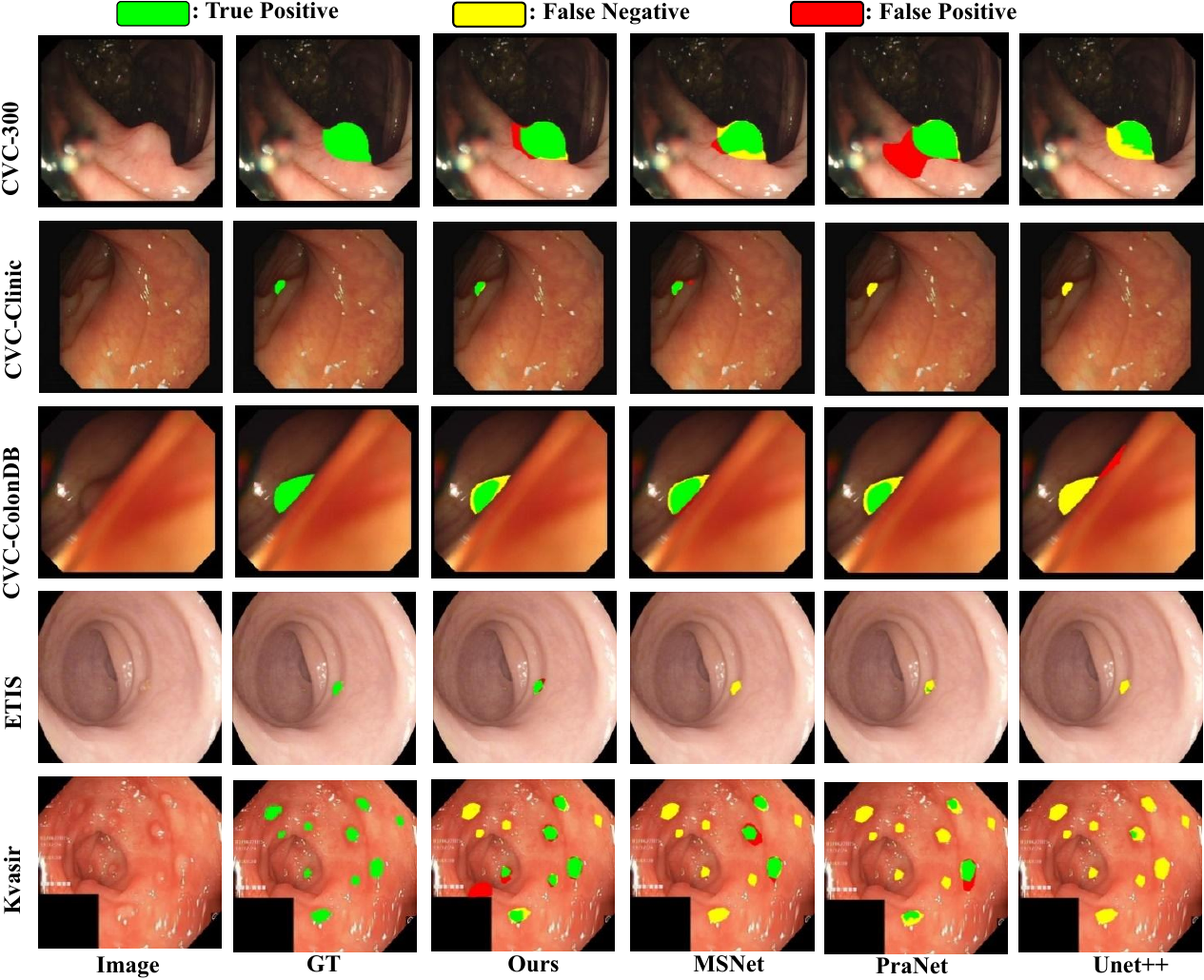}
\end{center}
   \caption{Visualization of the results of our model compared to other models. \textcolor{green}{Green} represents correct polyps, \textcolor{red}{red} are incorrect detections and \textcolor{yellow}{yellow} are missed polyps. 
   }
\label{fig:compare1}

\end{figure*}

\begin{figure*}[!ht]
\begin{center}
  \includegraphics[width=16cm,height=13cm]{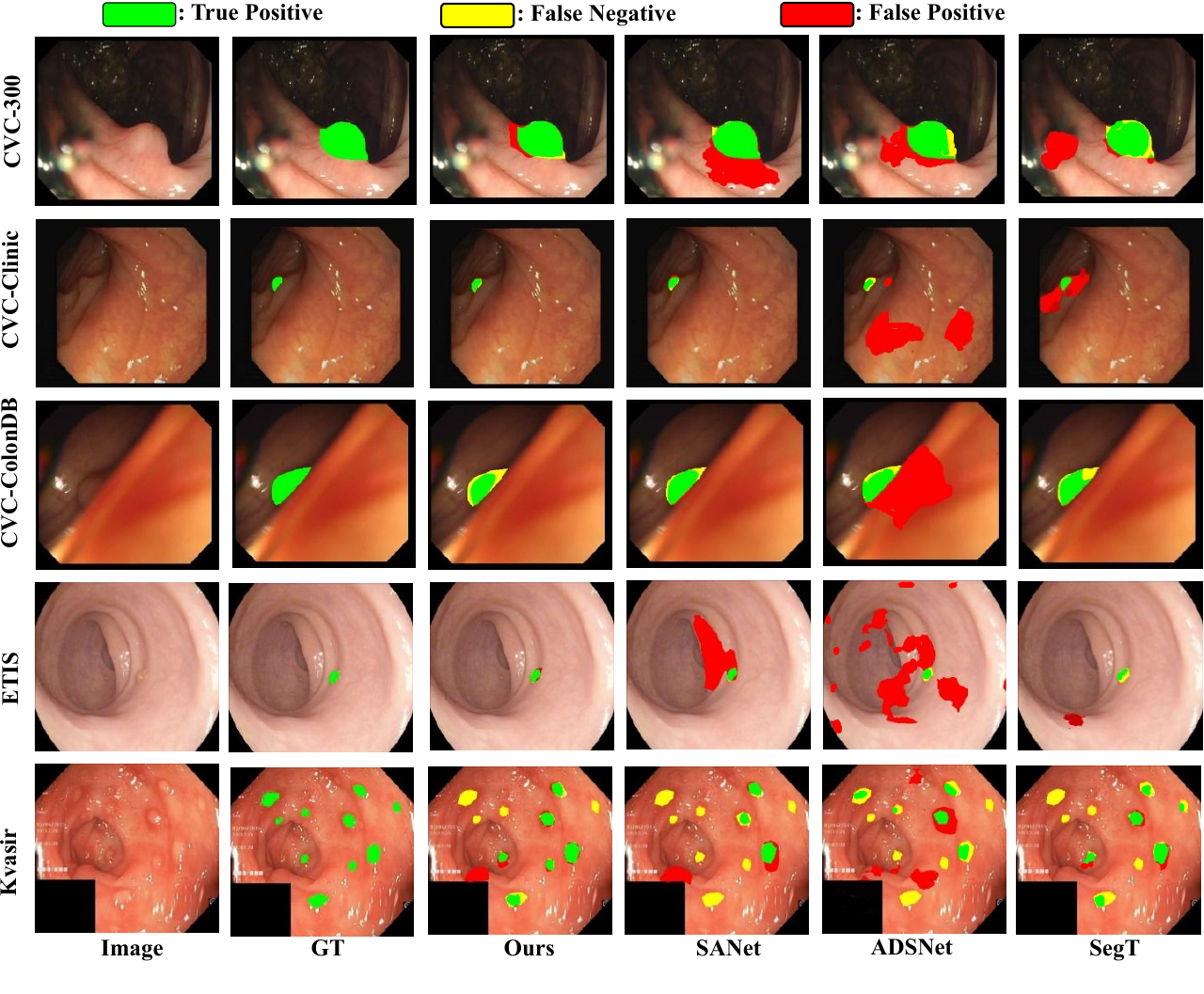}
\end{center}
   \caption{Visualization of the results of our model compared to other models.}
\label{fig:compare2}

\end{figure*}

\begin{figure*}[!ht]

\begin{center}

  \includegraphics[width=15cm,height=8cm]{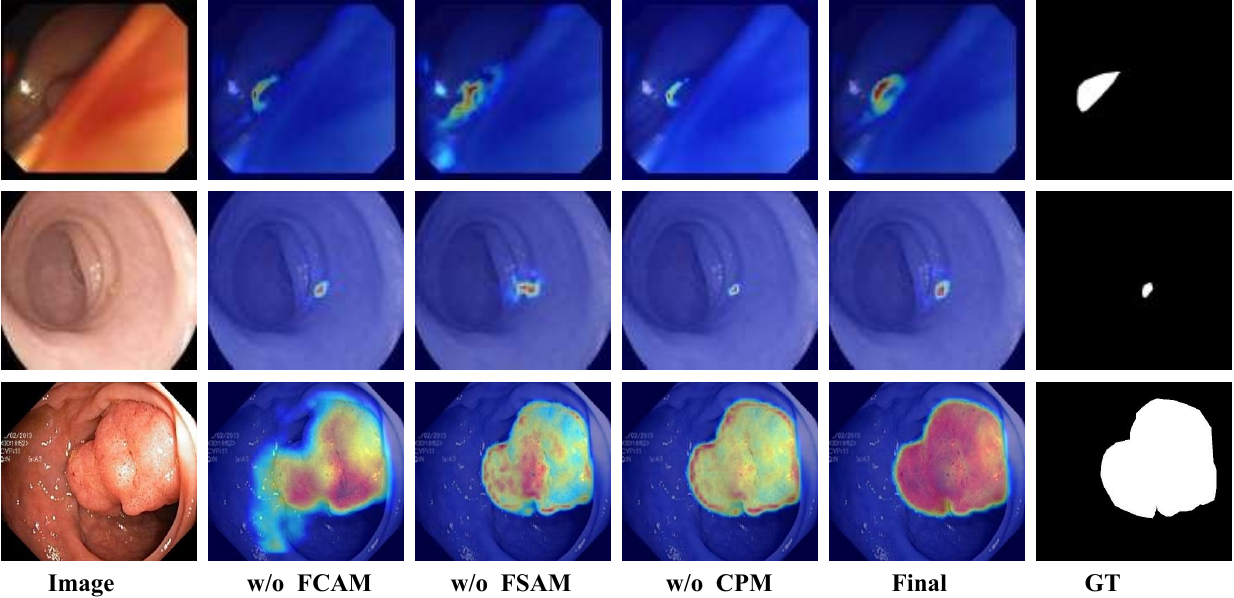}
\end{center}
   \caption{The ablation study results have been visually represented as heatmap. It is evident that the removal of any module leads to a substantial alteration in weighting, consequently resulting in the omission or incorrect detection of important elements.}
\label{fig:aba_cam}

\end{figure*}

\subsubsection{Visual Comparisons}
In Figs.\ref{fig:compare1} and\ref{fig:compare2}, we present the visualization results obtained from our PSTNet and eight other models, showcasing their performance on challenging examples. Our approach outperforms other methods in several critical aspects:
(a) Enhanced Detection of Small Polyps: Our method excels in capturing a more comprehensive range of small polyps. By effectively learning richer multi-scale information, it successfully identifies polyps of varying sizes, thus reducing the occurrence of missed detections.
(b) Improved Noise Suppression: Our model demonstrates superior noise suppression capabilities, effectively excluding polyps camouflaged by normal tissue. Given the frequent resemblance of polyps to their background, our approach leverages frequency domain information to facilitate the accurate identification of polyps amidst normal tissue.
(c) Enhanced Edge Prediction: The segmentation results generated by our model exhibit remarkable internal consistency and a closer alignment with ground truth data. This reflects our model's ability to predict edges more effectively, contributing to the overall accuracy and reliability of the segmentation results.

\subsection{Ablation Study}

An extensive ablation study was conducted, and all findings unequivocally affirmed the efficacy of each individual model component. The training, testing, and hyperparameter configurations precisely mirrored those elucidated in Section\ref{setting}. For illustration, Table\ref{tab:ablation} and Table~\ref{tab:abla_loss} show the ablation results for network structure components and loss functions, respectively.

\begin{table}[ht]
\centering
\caption{The table presents quantitative results from an ablation study, focusing on mDic and mIoU evaluation metrics. Various columns in the table indicate the influence of different configurations, with the best results highlighted in bold.}
\label{tab:ablation}
\renewcommand{\arraystretch}{1.1}
\setlength{\tabcolsep}{1.3mm}{
\begin{tabular}{c|c|cccc|c} 
\hline
Dataset                   & Metric(\%) & Bas. & w/o FCAM & w/o FSAM & w/o CPM & \textbf{Final}  \\ 
\hline
\multirow{2}{*}{Kvasir}   & mDic       & 91.2 & 92.5     & 91.9     & 92.3    & \textbf{93.5}   \\
                          & mIoU       & 86.0 & 87.8     & 86.5     & 87.3    & \textbf{89.5}   \\ 
\hline
\multirow{2}{*}{ClinicDB} & mDic       & 91.0 & 93.3     & 92.5     & 94.0    & \textbf{94.5}   \\
                          & mIoU       & 85.1 & 86.7     & 86.1     & 86.9   & \textbf{90.1}   \\ 
\hline
\multirow{2}{*}{CVC-300}  & mDic       & 87.3 & 90.1     & 88.7     & 90.5    & \textbf{91.0}   \\
                          & mIoU       & 80.0 & 81.3     & 80.6     & 82.2    & \textbf{84.7}   \\ 
\hline
\multirow{2}{*}{ColonDB}  & mDic       & 79.8 & 80.8     & 82.1     & 81.3    & \textbf{82.7}   \\
                          & mIoU       & 71.5 & 71.9     & 73.2     & 72.2    & \textbf{74.8}   \\ 
\hline
\multirow{2}{*}{ETIS}     & mDic       & 76.5 & 77.8     & 79.3     & 78.7    & \textbf{80.0}   \\
                          & mIoU       & 68.1 & 70.1     & 72.2     & 70.7    & \textbf{72.6}   \\
\hline
\end{tabular}}

\end{table}

\subsubsection{Network Components}
Our baseline (Bas.) is the shunted transformer\cite{ren2022shunted}, and we assess the effectiveness of the modules by removing or replacing components from the complete PSTNet and comparing the variants with the standard version. The standard version is denoted as ``PSTNet (ST+FCAM+FSAM+CPM)", where ``FCAM", ``FSAM" and ``CPM" indicate the usage of the FCAM, FSAM and CPM, respectively.

\textbf{Effectiveness of FCAM.} We investigate the contribution of the FCAM module. We trained a version of "PSTNet (w/o FCAM)". Table~\ref{tab:ablation} shows that the method without the FCAM module performs worse on the five datasets compared to the standard PSTNet. 
Notably, the mDic on the ClinicDB dataset drops by 1.2\%, and the mIoU drops by 3.4\%.
Meanwhile, the absence of FCAM is undeniably associated with a substantial introduction of noise (Fig.~\ref{fig:aba_cam}).

\textbf{Effectiveness of FSAM.} o analyze the effectiveness of FSAM, a version of "PSTNet (w/o FSAM)" is trained. Table~\ref{tab:ablation} shows that the mIoU metric drops on all datasets after removing the FSAM module, with the most significant drop observed on the CVC-300 dataset (from 84.7\% to 80.6\%). Meanwhile, the absence of FSAM is undeniably associated with a substantial introduction of noise (Fig.~\ref{fig:aba_cam}).

\textbf{Effectiveness of CPM.} To investigate the contribution of the CPM module to the model, we removed CPM from PSTNet and used the element summing operation as a substitute. This version of the training is called "PSTNet (w/o CPM)". Table~\ref{tab:ablation} shows that the standard version of PSTNet has better metrics than PSTNet (w/o CPM) on each dataset, with the most significant difference observed on the ClinicDB dataset ( mIoU: 90.1\% vs. 86.9\%). Fig.~\ref{fig:aba_cam} shows the benefits of CPM more intuitively. It is observed that the absence of CPM results in more pronounced errors in detail, and in some cases, even leads to missed detections.

\begin{table}[!ht]
\centering
\caption{The ablation study, focusing on the loss function, reports performance metrics as mDic(\%) and mIoU(\%) on the evaluation dataset. The best results are indicated in bold.}
\label{tab:abla_loss}
\renewcommand{\arraystretch}{1.1}
\begin{tabular}{cc|c|c|c}
\hline
Dataset                   & Metric(\%) & w/o (dice+focal) & w/o (wBCE) & \textbf{Final} \\ \hline
\multirow{2}{*}{Kvasir}   & mDic       & 92.1             & 92.2       & \textbf{93.5} \\
                          & mIoU       & 86.3             & 86.9       & \textbf{89.5} \\ \hline
\multirow{2}{*}{ClinicDB} & mDic       & 92.5             & 93.8       & \textbf{94.5} \\
                          & mIoU       & 87.1             & 89.3       & \textbf{90.1} \\ \hline
\multirow{2}{*}{CVC-300}  & mDic       & 89.8             & 90.6       & \textbf{91.0} \\
                          & mIoU       & 82.2             & 83.7       & \textbf{84.7} \\ \hline
\multirow{2}{*}{ColonDB}  & mDic       & 81.0             & 81.6       & \textbf{82.7} \\
                          & mIoU       & 72.5             & 73.9       & \textbf{74.8} \\ \hline
\multirow{2}{*}{ETIS}     & mDic       & 77.2             & 77.5       & \textbf{80.0} \\
                          & mIoU       & 70.2             & 72.2       & \textbf{72.6} \\ \hline
\end{tabular}

\end{table}
\subsubsection{Total Loss Function}
To assess the impact of each component of the loss function on model performance, we performed an ablation study. This study involved evaluating our model using three distinct loss function setups: 1) excluding the combined dice and focal loss ("w/o (dice+focal)"), 2) omitting the weighted binary cross-entropy loss ("w/o (wBCE)"), and 3) employing our ultimate combined loss function ("Final"). The results, presented in Table~\ref{tab:abla_loss}, indicate that our final loss function configuration persistently attains the highest mean Dice coefficient (mDic) and mean Intersection over Union (mIoU) scores across all datasets. This observation underscores the efficacy of our combined loss strategy in enhancing the robustness and accuracy of the segmentation outcomes.

\section{Conclusion}
In this work, we propose PSTNet, a novel approach tackling the challenges of low contrast, indistinct boundaries, and scale inconsistency in polyp segmentation through three key modules: FCAM for extracting frequency domain cues, FSAM for aligning and enhancing multi-scale features, and CPM for effectively combining them. Extensive experiments demonstrate PSTNet's superior performance over state-of-the-art models, highlighting the potential of incorporating frequency domain cues.

While focused on polyp segmentation, PSTNet's core concepts and modules can potentially extend to other medical imaging tasks with similar challenges, such as retinal imaging, lung nodule detection, and brain tumor segmentation. However, this requires careful consideration of modality-specific characteristics, network adjustments, feature extraction adaptations, and large-scale annotated datasets.

Future work will involve collaborating with domain experts, collecting relevant datasets, and validating our approach's effectiveness in new settings. We aim to explore advanced feature alignment techniques, incorporate domain-specific prior knowledge to enhance segmentation accuracy, and investigate integrating frequency domain cues with other state-of-the-art methods for broader medical image analysis applications. Overcoming challenges in data availability, computational efficiency, and model interpretability will be crucial for successful clinical translation.

\section{Acknowledgements}
This work was funded by National Natural Science Foundation of China (Nos. 62271074 and 82371962), Beiing Natural Science Foundation (L232133), CAMS
Innovation Fund for Medical Sciences (2022-I2M-C$\&$T-B-019) and National
High Level Hospital Clinical Research Funding (2022-PUMCH-033).

\bibliographystyle{IEEEtran}
\bibliography{ref}
\end{document}